%% file: root.tex
\documentclass[letterpaper, 10 pt, conference]{ieeeconf}  

\IEEEoverridecommandlockouts                              

\usepackage{amsmath} 
\usepackage{amssymb}  
\usepackage{booktabs}
\usepackage{bm}
\usepackage{algorithm}
\usepackage[noend]{algpseudocode}
\usepackage{lipsum}
\usepackage{graphicx}
\usepackage{subcaption}
\usepackage{xcolor}
\usepackage{cite}

\renewcommand{\baselinestretch}{0.96}

\title{\LARGE \bf
Interactive Robot Programming for Surface Finishing via Task-Centric Mixed Reality Interfaces}

\author{Christoph Willibald$^{1}$, Lugh Martensen$^{1,2}$, Thomas Eiband$^{1}$, and Dongheui Lee$^{1,3}$
\thanks{$^{1}$Institute of Robotics and Mechatronics, German Aerospace Center (DLR), Wessling, Germany.}%
\thanks{$^{2}$University of Lübeck, Lübeck, Germany.}%
\thanks{$^{3}$Autonomous Systems Lab, Institute of Computer Technology, TU Wien, Vienna, Austria.}%
\thanks{This work was supported in part by the German Federal Ministry of Education and Research (BMBF) through the “The Future of Value Creation — Research on Production, Services and Work” Program under Grant 02K20D032 and in part by the DLR internal Project “Factory of the Future Extended”}
}

\begin{document}

\maketitle
\thispagestyle{empty}
\pagestyle{empty}

\begin{abstract}
Lengthy setup processes that require robotics expertise remain a major barrier to deploying robots for tasks involving high product variability and small batch sizes. As a result, collaborative robots, despite their advanced sensing and control capabilities, are rarely used for surface finishing in small-scale craft and manufacturing settings. To address this gap, we propose a novel robot programming approach that enables non-experts to intuitively program robots through interactive, task-focused workflows. For that, we developed a new surface segmentation algorithm that incorporates human input to identify and refine workpiece regions for processing. Throughout the programming process, users receive continuous visual feedback on the robot's learned model, enabling them to iteratively refine the segmentation result. Based on the segmented surface model, a robot trajectory is generated to cover the desired processing area. We evaluated multiple interaction designs across two comprehensive user studies to derive an optimal interface that significantly reduces user workload, improves usability and enables effective task programming even for users with limited practical experience.

\end{abstract}

\input{sections/Introduction}
\input{sections/Related_Work}
\input{sections/Surface_Segmentation}

\input{sections/MR_Interface}
\input{sections/Evaluation_Segmentation}
\input{sections/User_Study}

\input{sections/Conclusion}

\addtolength{\textheight}{-12cm}   



\bibliographystyle{IEEEtran}
\bibliography{literature}

\end{document}

%% file: sections/Introduction.tex
\section{Introduction}
\label{sec:introduction}
Especially repetitive and physically intensive tasks are not attracting enough workers in today's competitive job market anymore \cite{feist2024imbalances}. Robotic solutions have the potential to alleviate the labour shortage problem by taking over unattractive tasks, while workers can focus on supervision and instruction. This provides a safe, ergonomic, and attractive work environment for the upcoming generations while making each workplace more productive. 
\begin{figure}[thpb]
    \centering
    \includegraphics[width=0.49\textwidth]{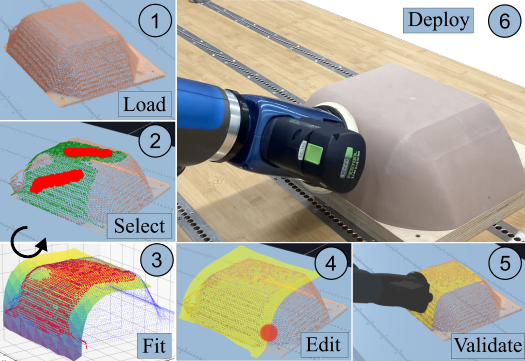}
    \caption{Our proposed human-robot interface for programming surface finishing tasks. The task-centric approach allows users to focus on the task to perform on the workpiece, instead of programming the robot's motions step-by-step.}
    \label{fig:Teaser}
    \vspace*{-0.5cm}
\end{figure}
To be economically viable in the context of small and medium-sized enterprises (SMEs), robotic solutions need to be highly re-configurable in minimal time to accommodate a variety of tasks \cite{perzylo2019smerobotics}. One particularly impactful yet challenging area for automation in small craft businesses is surface processing, including tasks such as sanding and polishing. These tasks are difficult to implement due to the significant variability in workpiece shapes and dimensions. Since collaborative robots have the required sensors and control algorithms to operate safely with humans in the same workspace and to reliably perform in-contact motions, they pave the way for such applications. However, the lack of flexible and intuitive software solutions for robot programming prevents SMEs from integrating collaborative robots into their workflow.

To improve this situation, active research in human-machine interface design and learning from demonstration aims to enable non-experts to naturally interact with robots to teach them new tasks \cite{makhataeva2020augmented, fu2023recent, wang2024eve}. Regardless of the essential role that humans play in this process, little focus has been put on measuring and improving the teacher's ability to provide good task instructions. Howard and Sena \cite{sena2020quantifying} show that there is a discrepancy between the user's belief of the robot's knowledge and its actual capabilities if the user receives insufficient feedback from the robot during the teaching process. This knowledge gap leads to a teaching performance degradation.

As shown in Fig.~\ref{fig:Teaser}, this work investigate the use of Mixed Reality (MR)-based human-robot interfaces to close the knowledge gap and to facilitate intuitive programming of surface finishing tasks.
Once the environment point cloud is loaded, the user selects the target area for processing and receives continuous visual feedback on the predicted surface model. The prediction can be adjusted and refined through additional user input until it aligns with the intended processing area. We introduce an algorithm that combines user input with a 3D point cloud to fit a surface model to the selected area (steps 2 and 3 in Fig.\ref{fig:Teaser}). The user can subsequently edit the surface model and virtually validate the robot’s finishing strategy before deploying it on the hardware. The contributions of this work are twofold

\begin{itemize}
    \item We developed a novel surface segmentation algorithm that uses an environment point cloud and user-demonstrated contact points to infer the shape primitive and the area of a workpiece to process. This enables the robot to utilize the optimal strategy for finishing the identified surface segment while eliminating the need to model the geometry of the workpiece.
    \item  We conducted comprehensive user studies to evaluate the usability and effectiveness of various human-robot interface concepts for programming of surface finishing tasks, leading to the derivation of optimal interaction patterns.
\end{itemize}

%% file: sections/Related_Work.tex
\section{Related Work}
\label{sec:related_work}
Robotic surface finishing typically begins with either locating the geometry of a known workpiece or generating a surface model for processing when no prior geometry information is available. In the latter case, several approaches \cite{wen2019novel, liang2023robotic, wang2021trajectory, alt2024robogrind} rely on obtaining a high-resolution point cloud of the workpiece, followed by statistical outlier filtering to eliminate scanning artifacts. Once the point cloud is refined, it is uniformly sliced by parallel intersection planes to generate a trajectory that covers the entire surface \cite{wen2019novel, liang2023robotic, wang2021trajectory} or to remove surface defects that were identified using curve fitting and outlier detection \cite{alt2024robogrind}. The points on these intersection planes, along with their surface normals, serve as via points for the robot's end-effector pose during the surface finishing process. To optimize these via points, Liang et al. \cite{liang2023robotic} apply the Douglas-Peucker algorithm \cite{douglas1973algorithms}, which reduces point density in areas of low curvature. Wang et al. \cite{wang2021trajectory} employ Non-Uniform Rational B-Spline (NURBS) curve fitting, ensuring smooth tool paths and consistent tool orientations throughout the trajectory. Xia et al. \cite{xia2023human} adopt a haptic exploration scheme to identify surface geometry and stiffness, where an operator guides the robot along the object’s surface while the robot maintains a constant contact force. The described approaches for unknown workpiece geometry still require great effort in manual preprocessing for the user. We reduce this effort by utilizing a model-fitting algorithm that is robust against outliers.


For scenarios with known workpiece geometries, several approaches \cite{liu2016industrial, sheng2001cad, liu2018region, schneyer2023segmentation, xiao2021model} utilize CAD models for surface segmentation and coverage planning. Liu et al. \cite{liu2016industrial} partition the CAD model into flat and curved segments, while \cite{sheng2001cad} use surface normals from tessellated CAD models to divide the surface into flat segments with different normal vector directions. In \cite{liu2018region}, the CAD model is uniformly sampled to generate a point cloud. Feature vectors comprising the point coordinates and their normal vectors are then used to segment the surface through k-means clustering. Schneyer et al. \cite{schneyer2023segmentation} consider both tool and workpiece geometry to identify segments that can be processed with the same strategy. Further, they adapt tool trajectories based on the contact area between the tool and the workpiece. Similarly, Wen et al. \cite{wen2022uniform} use a contact model incorporating applied normal force and local workpiece curvature to generate trajectories with adaptive spacing. In \cite{xu2024approach}, the point clouds of a finished and an unfinished workpiece are aligned with iterative closest point matching to detect uneven surfaces and to generate a sanding trajectory for their removal. In contrast to these approaches, our method does not rely on an a priori known geometry model of the workpiece. Instead, we perform surface segmentation directly on the point cloud, guided by human-demonstrated contact points. Our modified Random Sample Consensus (RANSAC) algorithm \cite{fischler1981random} uses the human input to reduce the necessary iterations until convergence to the correct surface model while providing immediate feedback to the user about the best model fit. Unlike the other methods, our approach gives users the control over the area to be processed, offering a flexible and intuitive solution for programming surface finishing tasks.

Mixed Reality (MR) interfaces in robotics have been used to simplify the collection of training data for learning from demonstration algorithms and to virtually evaluate the learned policies before deploying them on the hardware \cite{wang2024eve, dogangun2024rampa, hoang2022arviz}. In medical robotics, Augmented Reality (AR) is commonly used for preoperative planning and intraoperative navigation to reduce the mental load on surgeons and to improve the accuracy of surgical operations \cite{makhataeva2020augmented, fu2023recent, porpiglia2019three}. In industrial scenarios, 
MR interfaces were employed in welding applications for robotic path programming \cite{ong2020ar}, or to project an offline planned welding strategy onto objects \cite{tavares2019collaborative}. In the context of robotic surface finishing, Ciccarelli et al. \cite{ciccarelli2024advancing} validate the design of a robotic shoe-polishing workstation using Virtual Reality (VR) simulations to assess safety and ergonomy. Zhang \cite{zhang2023interactive} uses a head-mounted AR device to manually define and parameterize trajectories for surface polishing. In \cite{de2019intuitive}, users receive visual feedback of the applied contact force via a head-mounted AR device while demonstrating a surface finishing task leading to a more consistently applied force. However, results from a user study indicate that the feedback increases the mental load. This result is confirmed by \cite{quintero2018robot}, where kinesthetic teaching of contact tasks was found to be mentally less demanding than programming the same tasks using an AR interface. In contrast to this, a user study with industrial robot programmers \cite{stadler2016augmented} reports reduced mental load, but higher completion time when using an AR interface.
Our work goes beyond the state of the art in MR-assisted robot programming since our proposed interface hides the complexity of robot programming. With that, the user can focus on the task, i.e., \textit{what} to solve, and not on \textit{how} the robot is going to solve it. To balance between the mental load and the effectiveness of the interface, we systematically improved the interaction patterns and devices through the feedback from several user studies.

%% file: sections/Surface_Segmentation.tex
\section{Model Fitting and Surface Segmentation}
\label{sec:Surface_Segmentation}
Our proposed surface segmentation algorithm contains an iterative interaction approach with a user. First, the user selects contact points on a point cloud of the workpiece as a basis for fitting shape primitives.
Then, the resulting surface segment is displayed and can be continuously refined with additional contact points until it corresponds to the user's intended processing area.

\subsection{Background RANSAC Algorithm}
\label{sec:RANSAC}
RANSAC \cite{fischler1981random} is an iterative algorithm for fitting a model to data containing a significant proportion of outliers. It is commonly employed for visual odometry in robotics to find correspondences between features in two images \cite{he2020review}, or for detection of familiar geometric shapes in manipulation tasks \cite{martinez2022ransac}. The algorithm randomly selects a subset of data points, fits a model to them, and then separates the points into a set of inliers, that match the fitted model, and outliers that do not. This process is repeated multiple times to determine the model parameters yielding the highest inlier count. Unlike techniques like least squares, RANSAC starts with a minimal point set and expands it with consistent data, rather than optimizing for the best average fit. This makes it robust to outliers, which, in our case are points outside the target surface finishing region and artifacts from sensor noise.

The hyperparameters for RANSAC are: 1) the error threshold $\tau$ to determine whether a point is compatible with the model, 2) the maximum number of iterations, and 3) the minimum number of inliers required for a valid model, ensuring a sufficiently large consensus set to terminate the algorithm.

\subsection{Contact-Point-Guided Surface Segmentation}
\label{sec:Segmentation}
To fit a surface model with RANSAC, we need to solve two related problems:
1) finding the right model for the data, and 2) optimizing its free model parameters. 
Unlike classical RANSAC, our approach integrates user feedback into the model fitting process. Through a mixed reality (MR) interface and an interaction concept detailed in Sec.~\ref{sec:MR_Interface} and Fig.~\ref{fig:interactive}, users can specify contact points on the workpiece surface, which serve as input to the segmentation algorithm. Involving the user during the surface segmentation process offers several advantages: \newline
\textbf{A1)} Through the contact points, users guide the segmentation toward a specific workpiece area and refine the result online by adding more points. \newline
\textbf{A2)} These points limit the initial fitting region, reducing iterations needed for a good model fit, which enables online evaluation of several surface model candidates. \newline
\textbf{A3)} Continuous visual feedback helps users assess segmentation quality and stop the algorithm once satisfied, eliminating the need for RANSAC hyperparameters 2) and 3).
\vspace*{-0.2cm}
\begin{algorithm}
\caption{Contact Point Guided Surface Segmentation}\label{alg:CP-RANSAC}
\hspace*{\algorithmicindent}\!\textbf{Input:} Surface shape primitives $\mathcal{M}$, error threshold $\tau$\\
\hspace*{\algorithmicindent}\!\textbf{Output:}\ Best model $M^*$ with parameters $\theta^*$
\begin{algorithmic}[1]
\State Update $OP$
\While{user interaction active}
    \State Update $CP$ and sample a random subset $S \subset CP$
    \For{$M$ in $\mathcal{M}$}
        \State $\theta \gets \emptyset$, $OI \gets \emptyset$, $CI \gets \emptyset$
        \State Optimize model parameters $\theta$ to fit $S$
        \For{each data point $p_{obj} \in OP$}
            \If{error$(p_{obj}, M) < \tau$}
                \State Add $p_{obj}$ to $OI$
            \EndIf
        \EndFor
        \For{each data point $p_{con} \in CP$}
            \If{error$(p_{con}, M) < \tau$}
                \State Add $p_{con}$ to $CI$
            \EndIf
        \EndFor
        Compute $\varsigma^\theta_{M}(t)$ and $\varsigma^{\theta^*}_{M^*}(t)$
        \If{$\varsigma^\theta_{M}(t) > \varsigma^{\theta^*}_{M^*}(t)$}
            \State $\theta^* \gets$ Recompute model parameters using $OI$
            \State $M^* \gets M$
        \EndIf
    \EndFor
\EndWhile
\State \Return $M^*, \theta^*$
\end{algorithmic}
\end{algorithm}
\vspace*{-0.2cm}

Our proposed Algorithm~\ref{alg:CP-RANSAC} starts with obtaining a point cloud of the workspace containing the workpiece to process. We denote this point cloud as object points $OP$. After that, the interactive surface segmentation process starts. As illustrated in Fig.~\ref{fig:interactive}, the user selects or demonstrates contact points $CP$ in the intended area for surface processing through an interaction method. A random subset of points $S$ is sampled from $CP$. Next, the model parameters $\theta$ of surface model $M$ from a set of shape primitives $\mathcal{M}$, including planes, lines, spheres, and polynomial surfaces, 
are optimized for fitting the model to the points $S$. To fit polynomial surface models, a principal component analysis is first performed on the contact points. The two principal directions serve as the independent variables for fitting the polynomial surface model to the data points using least squares approximation. This approach allows users to select surfaces across all sides of the object, regardless of the workpiece's pose. Based on the fitted surface model $M$, the object and contact points $OP$ and $CP$ are divided into sets of outliers, whose distance from the surface model is above threshold $\tau$, and inliers $OI^\theta_{M}$ and $CI^\theta_{M}$, respectively. This result is used to compute a model score $\varsigma^\theta_{M}(t)$ for the current model $M$ with parameters $\theta$ according to

\vspace*{-0.4cm}
\begin{equation}
\label{eq:score}
    \varsigma^\theta_{M}(t) = \frac{\frac{|OI^\theta_{M}(t)|}{|OP(t)|}+\frac{|CI^\theta_{M}(t)|}{|CP(t)|}}{D_M}.
\end{equation}
Since the total number of contact points is significantly lower than the number of object points, the object and contact inlier counts $|OI^\theta_{M}(t)|$ and $|CI^\theta_{M}(t)|$ are normalized by the number of object and contact points in the model score. The process assumes that the user mainly selects contact points that lie within the intended area for surface processing and thus weights the contact points higher than the object points in the model score. The result is divided by the model dimension $D_M$ to penalize higher dimensional models. 
This follows the rationale that inliers of a simple model like a line are also inliers of a higher-dimensional model like a plane, however, if they can be explained by a simpler model, the simpler model is preferred. If the model score $\varsigma^\theta_{M}(t)$ is larger than the one of the best model $M^*$, the best model is updated and the parameters $\theta^*$ are recomputed to fit $M$ to all object inliers $OI$. Since the contact points change over time, the model score $\varsigma^{\theta^*}_{M^*}(t)$ of $M^*$ has to be recomputed at each iteration $t$. This process is iteratively repeated for all shape primitives in $\mathcal{M}$ until the user is satisfied with the segmentation result and terminates the process.

%% file: sections/MR_Interface.tex
\begin{figure}[thpb]
    \centering
    \begin{subfigure}{0.49\textwidth}
        \includegraphics[scale=0.98]{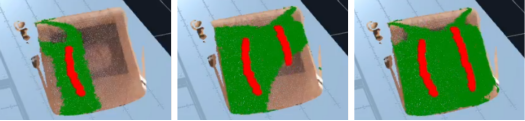}
        \caption{The user-selected contact points in red influence the segmentation result, which is visualized by highlighting the object inliers in green.}
        \label{fig:Chair}
    \end{subfigure}
    \begin{subfigure}{0.49\textwidth}
        \includegraphics[scale=0.98]{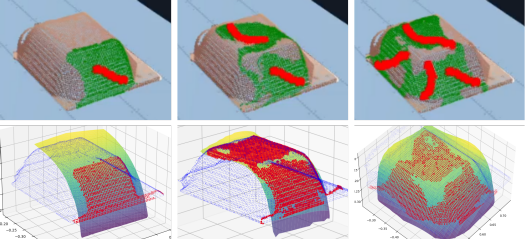}
        \caption{As shown in the lower row, the identified polynomial surface shape primitive adapts when new contact points are added.}
    \label{fig:Eval}
    \end{subfigure}
    \begin{subfigure}{0.49\textwidth}
        \includegraphics[scale=0.98]{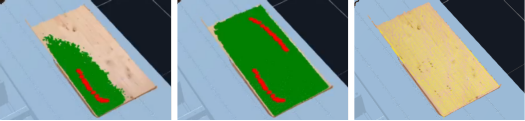}
        \caption{Guided by the continuous visual feedback, users can refine the segmentation result until it matches the intended processing region. The final surface model in yellow (right image) is based on the shape primitive and used to generate a robot processing trajectory.}
        \label{fig:Plane}
    \end{subfigure}
\caption{The iterative bidirectional interaction between user and robot illustrated with three different objects.}
\label{fig:interactive}
\vspace*{-0.5cm}
\end{figure}

\section{Mixed Reality Interface}
\label{sec:MR_Interface}
Bidirectional interaction between the robot and the user is a key aspect of our proposed approach. As shown in Fig.~\ref{fig:interactive}, users influence the segmentation by adding new contact points (red), while the system continuously informs the user about the learned task model by highlighting the object inliers $OI$ of the best model in a different color. The continuous feedback loop helps to minimize the gap between the user's belief of the robot's task model and the actual learned model, thereby reducing teaching failures and improving the accuracy of the learned task \cite{sena2020quantifying}. Several interface concepts are shown in Fig.~\ref{fig:Teaser},~\ref{fig:User_Study_1}, and~\ref{fig:User_Study_2}, where users can interact with the system through a combination of different MR technologies.

The key steps of programming the robotic surface finishing task are:
\begin{itemize}
    \item[S1)] Select or demonstrate contact points to specify the area for surface processing,
    \item[S2)] edit the area for processing by cropping the coverage to the desired region,
    \item[S3)] generate and virtually validate the trajectory for surface processing.
\end{itemize}

After checking the 3D environment point cloud containing the workpiece, the user provides contact points in one of two ways. The first method uses kinesthetic teaching, where the user moves the robot in gravity compensation mode across the surface to be processed while contact points between the sanding tool and the workpiece are recorded. Alternatively, the user manually selects points from the environment point cloud via an MR user interface, which serve as input for the contact point-guided surface segmentation. In both cases, the segmentation result is continuously visualized by highlighting the current model inliers in a different color. Based on the object inliers, a triangulated surface approximation is generated using Poisson surface reconstruction \cite{kazhdan2006poisson}. The Poisson reconstruction algorithm creates a smooth surface and can extrapolate into regions with low point density. The generated surface model is highlighted in yellow in Fig.~\ref{fig:Teaser} and~\ref{fig:User_Study_2}.  In step S2), the user manually removes parts of the yellow surface that should not be processed and the contour that encloses the desired surface finishing region is extracted. Using the identified surface model from S1) and the surface boundary from S2), a robot trajectory is generated to uniformly cover the entire segment. Following the approach in \cite{schneyer2023segmentation}, grid-based raster paths with a constant processing direction are used to ensure smooth tool motions. The optimal processing direction is parallel to the longest side of the smallest bounding rectangle of the segment, maximizing coverage line length and minimizing directional changes. Finally, the generated contact trajectory is visualized, and the robot virtually executes the trajectory before deploying the strategy on the real system.

%% file: sections/Evaluation_Segmentation.tex
\section{Qualitative Evaluation}
\label{sec:Experimental_Evaluation}

We evaluated our contact point-guided surface segmentation approach using the test object shown in Fig.~\ref{fig:Teaser} and~\ref{fig:Eval}, which measures 28×26×10~cm and is composed of various geometric shape primitives. As illustrated in Fig.~\ref{fig:Teaser} and~\ref{fig:Eval}, the segmentation result varies depending on the user-selected contact points. The set of candidate shape primitives $\mathcal{M}$ included planes, lines, spheres, second-order-, and third-order polynomial surfaces. In all evaluated cases, a third-order polynomial surface model achieved the highest model score $\varsigma^\theta_{M}$ and was identified as the optimal model $M^*$. As shown in Fig.~\ref{fig:Eval}, based on the same workpiece point cloud, the algorithm correctly identifies different target areas and surface model parameters depending on the selected contact points. These evaluation results demonstrate that the proposed segmentation algorithm effectively incorporates the user's intent.

%% file: sections/User_Study.tex
\vspace*{-0.3cm}
\section{User Studies}
\label{Sec:User_Studies}
To evaluate the effectiveness and usability of different interfaces and to iteratively derive the optimal user interaction concept, we conducted two user studies in which participants were asked to program two surface finishing tasks using several MR interfaces and interaction patterns. 

\subsection{Box Sanding}
\label{Sec:Box_Sanding}
\subsubsection{Setup and Task}
\label{Sec:Box_Setup}
\begin{figure}[thpb]
\vspace*{-0.9cm}
    \centering
    \includegraphics[trim={0 0 0 0}, clip, scale=0.96]{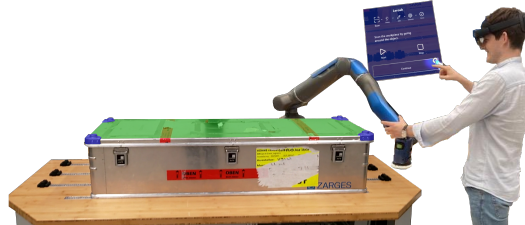}
    \caption{Setup of the box sanding task, where the upper plane of the box (marked in green) should be sanded by the robot, without processing the corner protectors highlighted in blue. As shown on the right, the participants can interact with the system via a head-mounted MR device and the robot in gravity compensation mode.}
    \label{fig:Box_setup}
\end{figure}
In the first user study, participants are tasked with programming the robot to sand the horizontal plane of the box shown in Fig.~\ref{fig:Box_setup}, while avoiding the corner protectors highlighted in blue. The box, measuring 160×45×45 cm, is placed on a table within the robot's workspace, ensuring that the entire green area is reachable. The environment point cloud is captured using a HoloLens 2, a head-mounted MR device that perceives spatial information and overlays holograms onto the real world. As shown in Fig.~\ref{fig:Box_setup}, a projected menu lets participants navigate through the different steps of the teaching procedure. Users can provide contact points in one of two ways: (1) by manually selecting points from the projected environment point cloud via the HoloLens interface, or (2) by demonstrating the surface processing strategy with the robot. In the latter approach, contact points are recorded when the sanding tool, mounted on the robot, makes contact with the workpiece. Contact points are identified if the normal force on the object exceeds a threshold of 3 N.
\subsubsection{Interfaces and Human-Robot Interaction}
\label{Sec:Box_Procedure} 
\begin{figure}[thpb]
    \vspace*{-0.2cm}
    \centering
    \includegraphics[trim={0 0 0 0}, clip, scale=0.3]{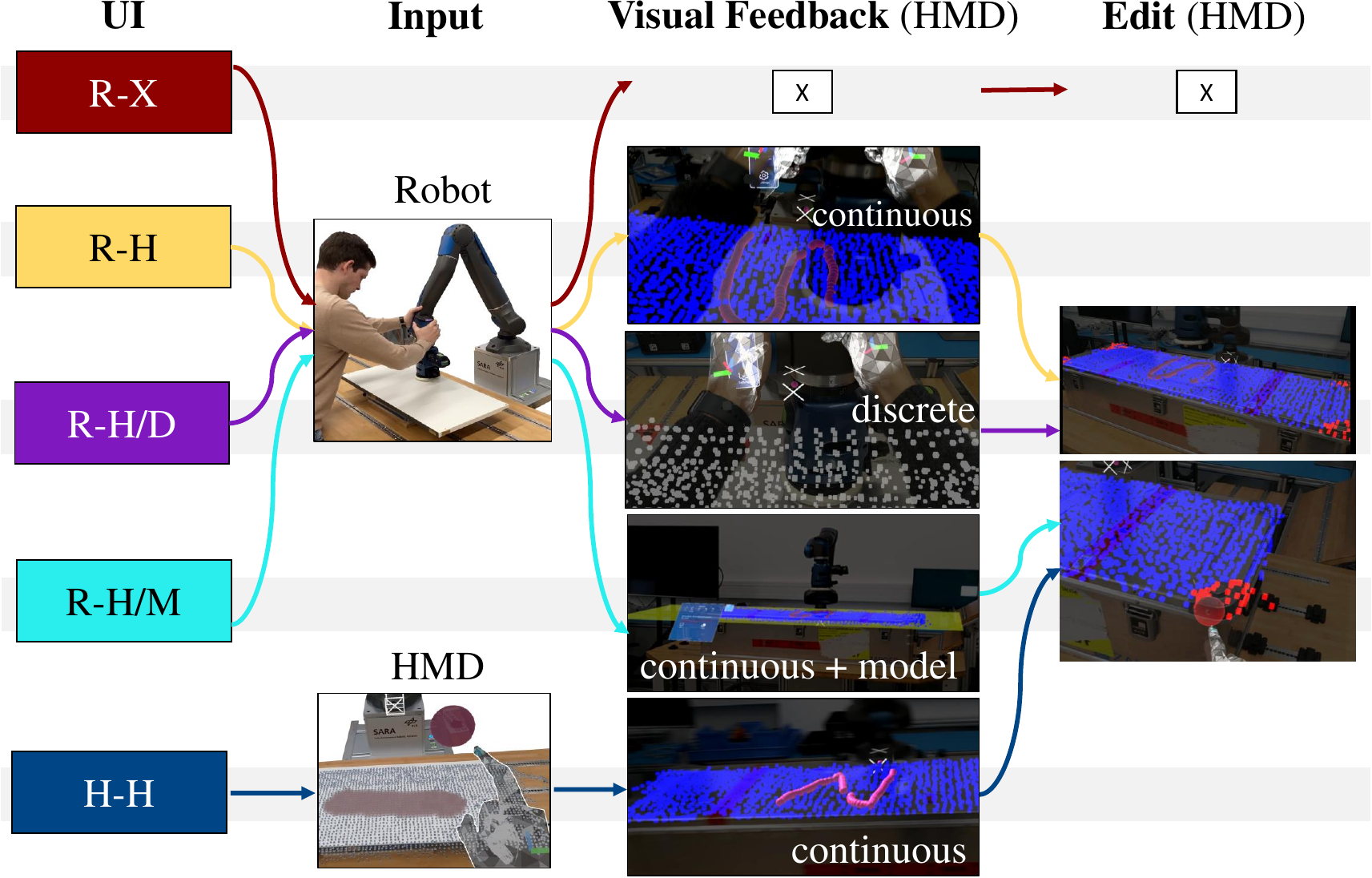}
    \caption{The five user interface designs compared in the box sanding study. In R-X, no feedback is provided and the participants can not edit the segmentation result after the demonstration. The other interfaces highlight object inliers in blue and contact points in pink. In the edit phase, areas to exclude from processing can be selected through the HMD.}
    \label{fig:User_Study_1}
    \vspace*{-0.5cm}
\end{figure}
 \begin{table}[thpb]
    \caption{User interfaces compared in the box sanding (1) and chair sanding (2) user studies.}
    \label{tab:Interfaces_study1}
    \centering
    \begin{tabular}{l|l|l|c}
        \toprule
        Interface & Input & Visual Feedback & Study\\
        \midrule
        R-H/D & Robot & HMD (Discrete) & 1\\
        R-H/M & Robot & HMD (+ Model) & 1\\
        R-X &  Robot & None & 1+2\\
        R-H & Robot & HMD & 1+2\\
        H-H & HMD & HMD & 1+2\\
        R-T & Robot & Tablet \& Projector & 2\\
        T-T & Tablet & Tablet \& Projector & 2\\
         \bottomrule
    \end{tabular}
\end{table}

We compare five different interface designs shown in Fig.~\ref{fig:User_Study_1} and summarized in Table~\ref{tab:Interfaces_study1}. In the first four setups, participants provide contact points by guiding the robot in gravity compensation mode, whereas in the H-H interface, users select contact points manually via the head-mounted device (HMD). During contact point demonstration, the R-H, R-H/M, and H-H interfaces continuously update and project contact points and object inliers identified with our proposed segmentation approach onto the workpiece through the HMD. The R-H/M interface additionally projects a model of the identified shape primitive into the workspace. The R-H/D interface provides visual feedback about the learned model through the HMD only after the user completes the contact point demonstration phase. In all interfaces except R-X, participants can use the HMD to mark areas within the identified segment that should not be processed by the robot. Since the R-X interface does not provide any feedback and does not allow to interact with the learned model, the last step is skipped for this condition.

\subsubsection{Study Design}
\label{Sec:Box_Study_Design}
The user study involved 20 participants (16 male and four female) with a mean age of $25.05 \pm{3.56}$~years (ranging from 19 to 35). Eight participants had proficient knowledge in robotics, but none had prior professional experience with sanding.

At the start of the study, participants were briefed on the study’s objectives and procedure followed by 10 minutes of familiarization with the HMD and practicing moving the robot in gravity compensation mode. Following this, they programmed the surface finishing task using all five user interfaces, where the task order among the participants was permuted using a Latin square design. After using each interface, participants rated their understanding of the learned model on various aspects using a 5-point Likert-type scale. They also completed the NASA TLX questionnaire \cite{hart1988development} to assess task load and the Questionnaire for Measuring the Subjective Consequences of Intuitive Use (QUESI) \cite{hurtienne2010quesi} to evaluate interface usability. Finally, a guided interview was conducted to collect general feedback. For questionnaire items, a repeated measures ANOVA was conducted. In case of violation of sphericity (Mauchly’s sphericity test), Huynh-Feldt $(\varepsilon > .75)$ or Greenhouse-Geisser $(\varepsilon < .75)$ corrections were applied. Post-hoc tests with Bonferroni correction were performed to identify significant differences between interfaces.

\subsubsection{Hypotheses}
\label{Sec:BOX_Hypotheses}
\begin{itemize}
    \item Visual feedback via an HMD:
    \begin{itemize}
        \item reduces the workload during programming ($\mathcal{H}1.1$)
        \item increases the users' model understanding ($\mathcal{H}1.2$)
    \end{itemize}  
    \item Selection of contact points by hand:
    \begin{itemize}
        \item reduces the workload during programming ($\mathcal{H}2.1$)
        \item increases usability and user satisfaction ($\mathcal{H}2.2$)
    \end{itemize}
    \item Continuous feedback increases the users' model understanding compared to discrete feedback ($\mathcal{H}3$)
\end{itemize}

\subsubsection{Quantitative Results}
\label{Sec:Box_Results}
\begin{figure}[thpb]
\centering
\begin{subfigure}[t]{.25\textwidth}
  \centering
  \includegraphics[scale=0.165]{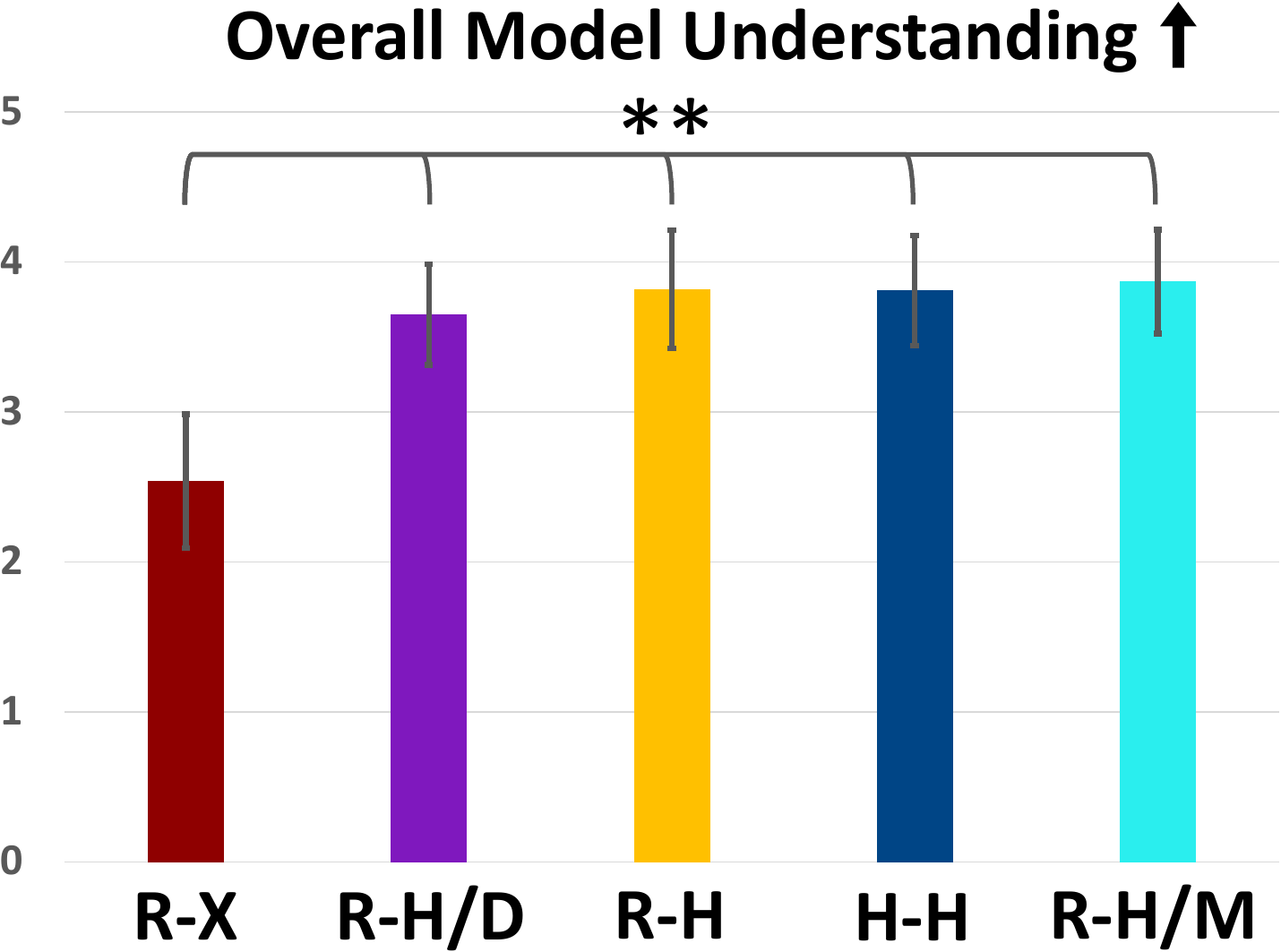}
  \label{fig:sub11}
\end{subfigure}%
\begin{subfigure}[t]{.25\textwidth}
  \centering
  \includegraphics[scale=0.165]{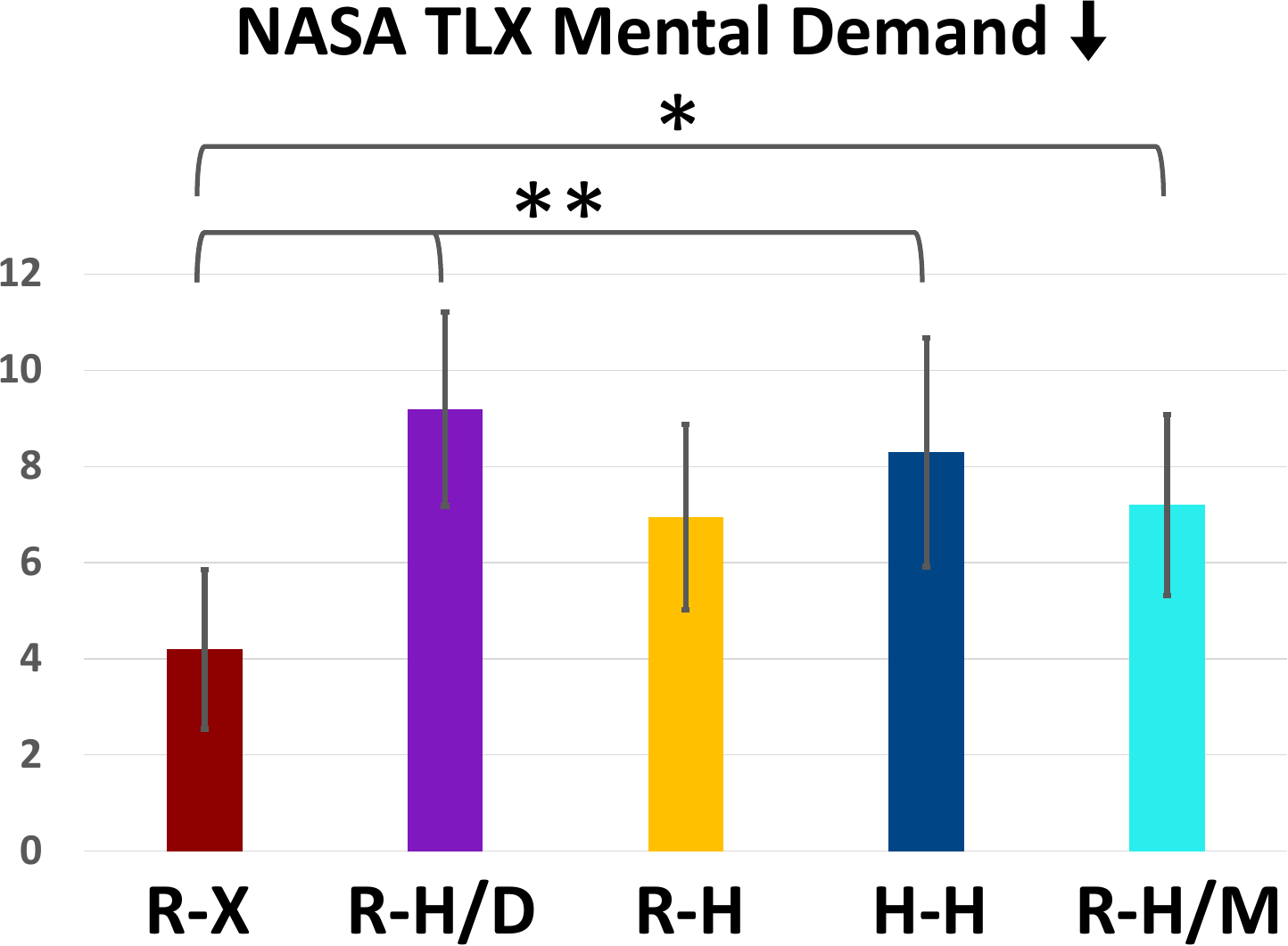}
  \label{fig:sub12}
\end{subfigure}
\caption{The overall model understanding score and the NASA TLX mental demand for the interfaces of the box sanding user study. Error bars indicate 95\% confidence intervals, statistical significance level $\bm{*}: p < 0.05$, $\bm{**}: p < 0.001$.}
\label{fig:Box_Model_NASA_TLX}
\vspace*{-0.5cm}
\end{figure}

The NASA TLX questionnaire indicated a significant ANOVA main effect for the categories mental demand $F(4,76)\!=\!6.85, p\!<\!.001$, and physical demand $F(4,76)\!=\!7.677, p\!<\!.001$. As shown in Fig.~\ref{fig:Box_Model_NASA_TLX}, post-hoc comparisons revealed that mental demand was significantly lower for R-X compared to R-H/D, H-H, and R-H/M . However, no significant difference was observed between R-X and R-H. Physical demand was significantly lower in the H-H condition compared to all other interfaces, with $p\!<\!.001$ for R-X, R-H, R-H/D, and $p\!<\!.05$ for R-H/M.

To evaluate several aspects of participants' model understanding, we asked: was it clear (Q1) what task model was learned, (Q2) when to stop the demonstration, (Q3) how to influence the model with your actions, (Q4) how the robot is learning, and (Q5) what information the robot needs to build a model. For all questions, as well as for overall model understanding, a repeated measures ANOVA revealed a significant main effect $(p\!<\!.001)$. Post-hoc tests showed that the R-X condition was consistently rated lower across all categories compared to the other conditions, with $p\!<\!.001$ for all categories (see Fig.~\ref{fig:Box_Model_NASA_TLX} for overall score) except for Q5, where the significance level was $p\!<\!.05$.
\subsubsection{Discussion}
\label{Sec:Box_Discussion}
The first hypothesis, $\mathcal{H}1.1$, could not be confirmed by the study results. Although the R-H interface achieved the best (not statistically significant) scores in the performance and frustration categories of the NASA TLX and the goal achievement category of the QUESI questionnaire, the NASA TLX results also indicate an increased mental workload when visual feedback is provided through the HMD. Guided user interviews support this finding, indicating that navigating the application via the HMD, its limited field of view, and the dynamically changing visualization of the segmentation results made it difficult to interpret and process all relevant information. Hypothesis $\mathcal{H}1.2$ was confirmed with regard to all aspects of user model understanding.

It could not be confirmed that manually selecting contact points via the HMD reduces the workload ($\mathcal{H}2.1$) or increases usability and satisfaction ($\mathcal{H}2.2$). However, the physical demand subcategory of the NASA TLX did show a significant reduction when participants demonstrated contact points via the HMD instead of physically guiding the robot. Additionally, although the results from the QUESI questionnaire were not statistically significant, they indicate that the H-H interface felt most familiar to participants and required the least amount of learning effort.

Finally, while the R-H/D interface with discrete feedback consistently scored lower than interfaces providing continuous feedback in all model understanding categories, these differences were not statistically significant, and thus, hypothesis $\mathcal{H}3$ could not be confirmed.
\vspace*{-0.1cm}
\subsection{Chair Sanding}
\label{Sec:Chair_Sanding}

\begin{figure*}[thpb]
    \centering
    \includegraphics[trim={0 0 0 0}, clip, scale=0.52]{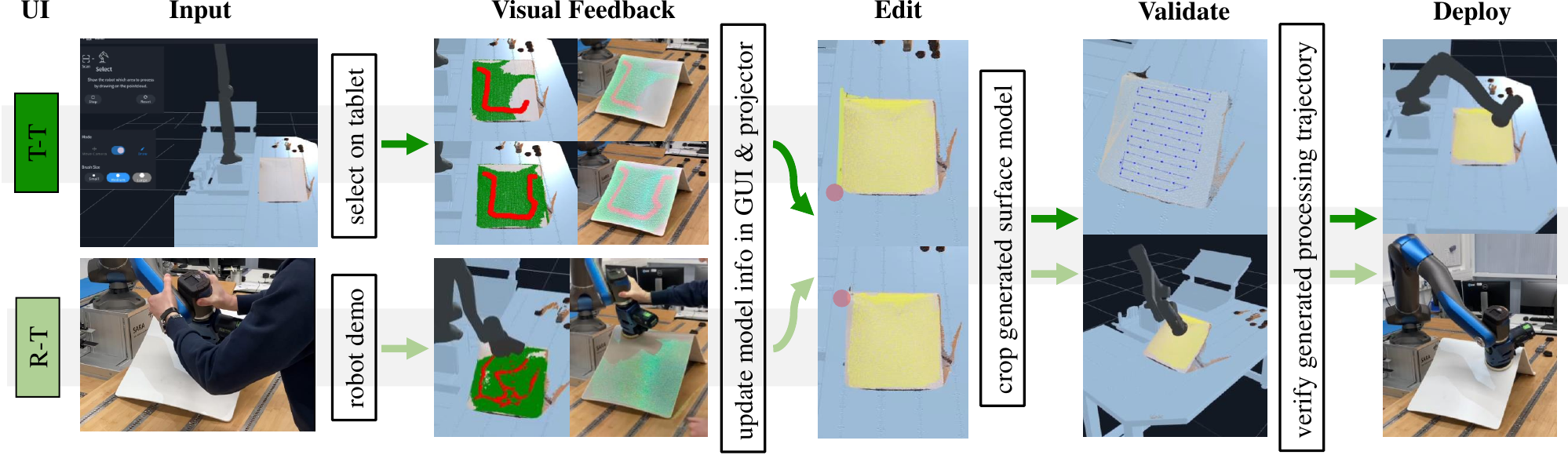}
    \caption{The two new user interfaces T-T and R-T. In T-T, the user selects contact points via the GUI on the tablet, while participants demonstrate the sanding strategy with the robot in R-T. In both cases, information about the learned model is provided via the tablet, and the projector that augments the information directly onto the workpiece. Participants can edit the final surface model and validate the generated robot trajectory on the tablet before deploying the strategy on the robot.}
    \label{fig:User_Study_2}
    \vspace*{-0.5cm}
\end{figure*}

Building on the findings of the first user study and participants’ feedback, we revised the interface concepts accordingly. Particularly, taking into account the mentioned limitations of using an HMD, we developed a new MR interface shown in Fig.~\ref{fig:User_Study_2} that combines a tablet-based graphical user interface (GUI) with a projector. The projector overlays model information directly onto the surface of the workpiece, eliminating the need for an HMD.
\begin{figure}[thpb]
    \centering
    \includegraphics[trim={0 0 0 0}, clip, scale=0.96]{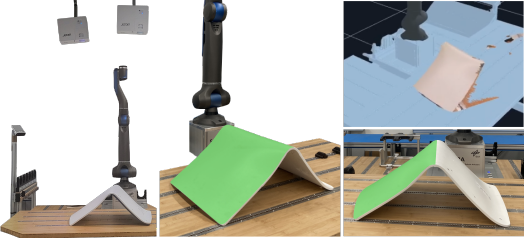}
    \caption{Setup of the chair sanding task, where only the area of the chair highlighted in green should be sanded by the robot. The 3D point cloud recorded by the camera on the left is displayed in a GUI shown in the upper right. With the projectors mounted above the robot, additional information can be displayed directly onto the surface of the chair.}
    \label{fig:Chair_setup}
    \vspace*{-0.5cm}
\end{figure}

\subsubsection{Setup and Task}
\label{Sec:Chair_Setup}
In this user study, participants have to program the robot to sand the chair's surface area highlighted in green in Fig.~\ref{fig:Chair_setup}. The target area has varying curvatures along both principal directions, adding complexity to the task compared to the first user study. The chair, measuring 65x40x25~cm, is positioned within the robot's workspace ensuring that the entire green area is reachable by the robot and visible by the camera used to capture the environment point cloud. In addition to the two contact point demonstration strategies from the first study, participants can now also manually select contact points from the environment point cloud using a tablet-based GUI. 
\subsubsection{Interfaces and Human-Robot Interaction}
\label{Sec:Chair_Interfaces_Interaction}
In addition to the R-X, H-H, and R-H interfaces from the box sanding study, two new interface design are tested, leading to a total of five compared interfaces. The two new interface designs R-T and T-T, shown in Fig.~\ref{fig:User_Study_2}, combine a tablet-based GUI with a projector-based overlay of model information onto the workpiece surface. Contact points are selected via the tablet-based GUI in T-T, and by robot demonstration in R-T. After providing contact points, participants can edit the final area for processing through the GUI. In the last step, users can virtually validate the generated trajectory before deploying the strategy on the robot. 
\subsubsection{Study Design}
\label{Sec:Chair_Study_Design}
The user study involved 20 participants (17 male and three female) with a mean age of $25.35 \pm{2.98}$~years (ranging from 20 to 32). Seven participants had proficient knowledge in robotics, but none had prior professional experience with sanding.

The study followed the same procedure and statistical analysis as the box sanding study described in Sec.~\ref{Sec:Box_Study_Design}, with two differences. First, before using each interface to program the task, participants watched a one-minute introductory video explaining its functionality and interaction methods. Second, in addition to the NASA TLX and QUESI questionnaires, participants completed the User Experience Questionnaire (UEQ) \cite{laugwitz2008construction} after using each interface to assess aspects such as attractiveness, efficiency, and stimulation.
\subsubsection{Hypotheses}
\label{Sec:Chair_Hypotheses}
\begin{itemize}
    \item The T-T interface design:
    \begin{itemize}
        \item causes least workload during programming ($\mathcal{H}1$)
        \item increases the users' model understanding without increasing the cognitive load ($\mathcal{H}2$)
        \item is the most user friendly ($\mathcal{H}3$)
        \item is most effective to fulfill the task goal ($\mathcal{H}4$)
    \end{itemize}
\end{itemize}
\subsubsection{Quantitative Results}
\label{Sec:Chair_Results}
A repeated measures ANOVA revealed statistically significant differences across interfaces for the overall NASA TLX score $F(2.35,44.65)\!=\!12.75,p\!<\!.001$, as well as for the subcategories frustration $F(4,76)\!=\!9.19,p\!<\!.001$, and mental demand $F(4,76)\!=\!10.68,p\!<\!.001$. Significant main effects were also found in the QUESI results for the overall score $F(4,76)\!=\!13.66,p\!<\!.001$, and the subcategories mental workload $F(4,76)\!=\!14.64,p\!<\!.001$, goal achievement $F(2.71,51.5)\!=\!5.59,p\!<\!.005$, and perceived error rate $F(4,76)\!=\!8.87,p\!<\!.001$. For the UEQ, a significant main effect was found for the efficiency subcategory $F(4,76)\!=\!9.27,p\!<\!.001$.

\begin{figure}
\centering
\begin{subfigure}[t]{.25\textwidth}
  \centering
  \includegraphics[scale=0.15]{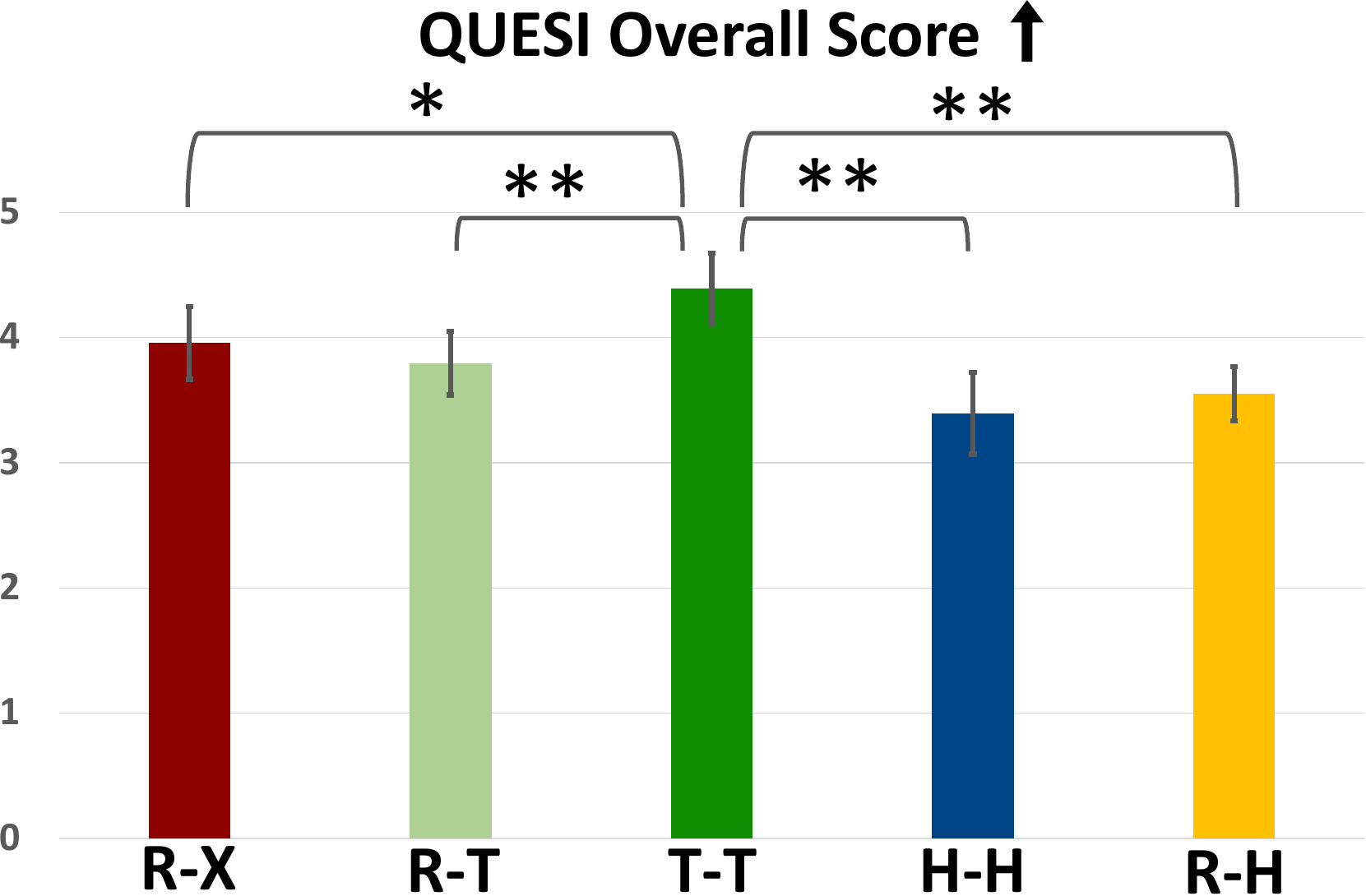}
  \label{fig:sub1}
\end{subfigure}%
\begin{subfigure}[t]{.25\textwidth}
  \centering
  \includegraphics[scale=0.15]{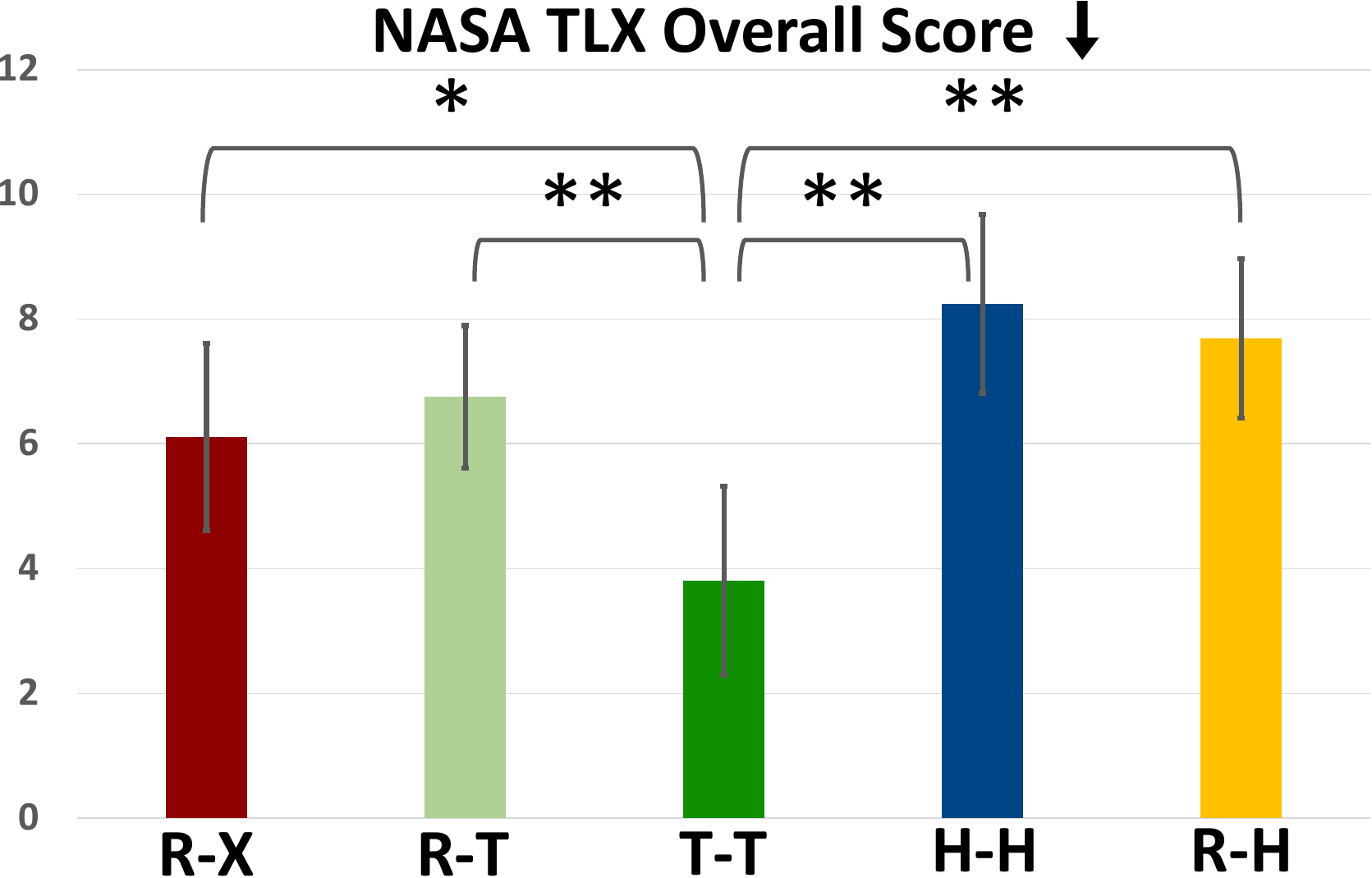}
  \label{fig:sub2}
\end{subfigure}
\caption{The overall QUESI and NASA TLX scores for the interfaces of the chair sanding user study. T-T outperforms all other interfaces in both QUESI and NASA TLX overall scores. Error bars indicate 95\% confidence intervals, statistical significance level $\bm{*}: p < 0.05$, $\bm{**}: p < 0.001$.}
\label{fig:Chair_QUESI_TLX_Overall}
\vspace*{-0.5cm}
\end{figure}

As shown in Fig.~\ref{fig:Chair_QUESI_TLX_Overall}, post-hoc comparisons demonstrated that the newly introduced T-T interface  significant outperformed all others in terms of both NASA TLX and QUESI overall scores. Specifically, in the frustration subcategory of NASA TLX and the perceived error rate subcategory of QUESI, T-T was rated significantly better than all HMD-based interfaces $(p\!<\!.001)$. Frustration was also significantly lower for T-T compared to R-T $(p\!<\!.05)$, though no significant differences were found between T-T and R-X in terms of frustration or error rate.

In terms of mental workload, both T-T and R-X received significantly better ratings compared to the remaining interfaces. QUESI mental workload post-hoc tests confirmed this, showing $p\!<\!.001$ for T-T and $p\!<\!.05$ for R-X, with the exception of the R-X vs. R-T comparison, which did not reach significance. For the NASA TLX mental demand subcategory, both R-X and T-T showed significantly reduced scores compared to H-H $(p\!<\!.001)$, and to R-H and R-T $(p\!<\!.05)$. Moreover, T-T demonstrated significantly improved efficiency (UEQ) and goal achievement (QUESI) ratings compared to R-X $(p\!<\!.05)$ and all HMD-based interfaces $(p\!<\!.001)$.

Finally, the findings from the box sanding study regarding participants’ model understanding were confirmed. R-X was consistently rated significantly lower across all model understanding questions and the overall understanding score compared to all other interfaces.
\subsubsection{Discussion}
\label{Sec:Chair_Discussion}
The results of the user study confirmed all hypotheses $\mathcal{H}1$ - $\mathcal{H}4$. The overall NASA TLX scores support $\mathcal{H}1$, indicating that the T-T interface imposes the lowest workload on users during the programming process. Further analysis of QUESI and TLX subcategories reveals that the mental demand associated with T-T can be reduced to a level comparable to R-X. Combined with the superior task model understanding achieved with T-T compared to R-X, these findings demonstrate that T-T enhances task model comprehension without increasing cognitive load ($\mathcal{H}2$).

\begin{figure}[thpb]
    \vspace*{-0.2cm}
    \centering
    \includegraphics[scale=0.22]{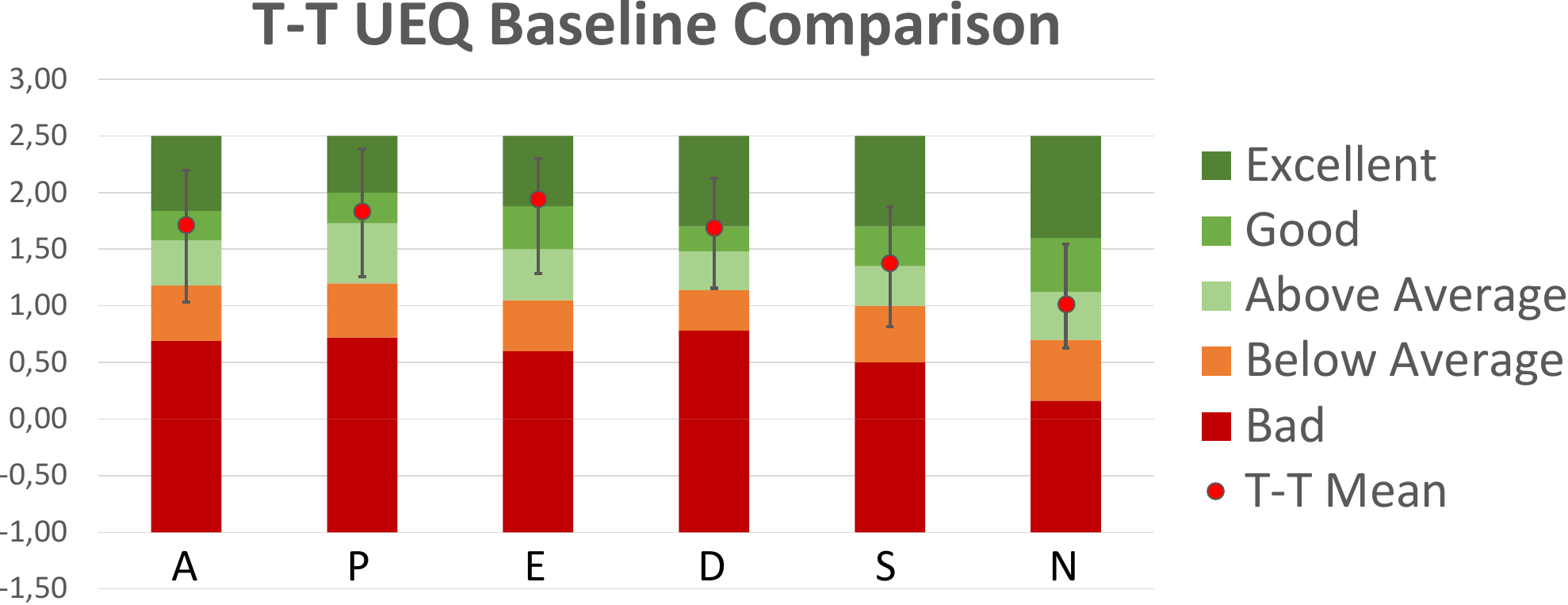}
    \caption{Benchmarking result of the T-T interface compared to 452 interfaces in the UEQ scales Attractiveness (A), Perspicuity (P), Efficiency (E), Dependability (D), Stimulation (S), and Novelty (N). The category excellent includes the top 10\% interfaces, good: top 25\%, above average: top 50\%.}
    \label{fig:T-T_C_UEQ}
    \vspace*{-0.5cm}
\end{figure}

The statistically significant improvement in the overall QUESI score confirms $\mathcal{H}3$, stating that T-T is the most intuitive interface for programming robotic surface finishing tasks. Further, T-T reduces the error rate and frustration compared to interfaces requiring an HMD, supporting the claim that several pieces of information in a limited field of view hinders effective interpretation and processing. The UEQ benchmarking results for the T-T interface in Fig.~\ref{fig:T-T_C_UEQ} further reinforce $\mathcal{H}3$. Compared to 452 UEQ evaluations of diverse interfaces such as business applications, web shops, or mobile apps \cite{schrepp2017construction}, T-T ranks among the top 25\% of interfaces across all UEQ dimensions except for novelty, in which it is still above average. Moreover, T-T outperforms all other interfaces evaluated in our study across all UEQ scales, except novelty and stimulation, where HMD-based interfaces received higher ratings.

In UEQ's efficiency category, T-T ranks within the top 10\% of the benchmark dataset and scores significantly better than R-X and the HMD-based interfaces. The same results can be shown for the QUESI goal achievement category, further validating $\mathcal{H}4$. While tablet-based interfaces like T-T may be perceived as less novel or stimulating than HMD-based alternatives, they are highly efficient and effective in achieving the task goals.

%% file: sections/Conclusion.tex
\section{Conclusion}
\label{sec:Conclusion}

We presented a novel surface segmentation algorithm designed to facilitate task-centric programming of robotic surface finishing tasks. By incorporating human feedback, the algorithm enables intuitive selection and refinement of workpiece regions for processing. Through bidirectional interaction, where users iteratively guide the segmentation with additional input while receiving real-time feedback on the learned task model, users gain a significantly improved understanding of the robot's internal task model. This improves their ability to provide effective instructions.

Across two user studies, we evaluated multiple human-robot interface designs. Our findings regarding increased mental demand of HMDs confirmed the user study results from \cite{de2019intuitive}, \cite{quintero2018robot}, and \cite{stadler2016augmented}, which led to the development of an optimal interface configuration: a combination of a tablet-based GUI for selecting contact points and receiving model feedback, and a projector for augmenting model information directly onto the workpiece in the robot's workspace. This setup eliminates the need for an HMD, addressing usability and mental load concerns related to limited field of view and interface complexity. The combined tablet-projector interface consistently outperformed alternative designs in terms of usability, workload reduction, and effectiveness. Notably, it also enabled participants with limited experience in robotics and no experience in sanding to successfully program a new surface finishing task in just a few minutes, highlighting the potential of our approach for intuitive robot programming.

%% file: literature.bib
@book{feist2024imbalances,
  title={Imbalances between supply and demand: Recent causes of labour shortages in advanced economies},
  author={Feist, Lisa},
  year={2024},
  publisher={ILO Working Paper}
}

@article{perzylo2019smerobotics,
  title={SMErobotics: Smart robots for flexible manufacturing},
  author={Perzylo, Alexander and others},
  journal={IEEE RA-M},
  volume={26},
  year={2019},
  publisher={IEEE}
}

@article{sena2020quantifying,
  title={Quantifying teaching behavior in robot learning from demonstration},
  author={Sena, Aran and Howard, Matthew},
  journal={The International Journal of Robotics Research},
  volume={39},
  year={2020},
  publisher={SAGE Publications Sage UK: London, England}
}

@article{fischler1981random,
  title={Random sample consensus: a paradigm for model fitting with applications to image analysis and automated cartography},
  author={Fischler, Martin A and Bolles, Robert C},
  journal={Communications of the ACM},
  volume={24},
  year={1981},
  publisher={ACM New York, NY, USA}
}

@inproceedings{alt2024robogrind,
  title={RoboGrind: Intuitive and Interactive Surface Treatment with Industrial Robots},
  author={Alt, Benjamin and others},
  booktitle={ICRA},
  year={2024},
  organization={IEEE}
}

@inproceedings{wen2019novel,
  title={A novel robotic system for finishing of freeform surfaces},
  author={Wen, Yalun and others},
  booktitle={ICRA},
  year={2019},
  organization={IEEE}
}

@article{liang2023robotic,
  title={A robotic polishing trajectory planning method for TBCs of aero-engine turbine blade using measured point cloud},
  author={Liang, Xufeng and others},
  journal={Industrial Robot: the international journal of robotics research and application},
  volume={50},
  year={2023},
  publisher={Emerald Publishing Limited}
}

@article{douglas1973algorithms,
  title={Algorithms for the reduction of the number of points required to represent a digitized line or its caricature},
  author={Douglas, David H and Peucker, Thomas K},
  journal={Cartographica: the international journal for geographic information and geovisualization},
  volume={10},
  year={1973},
  publisher={University of Toronto Press}
}

@inproceedings{xu2024approach,
  title={An Approach for Precise Robotic Sanding Applications of Uneven Surfaces Based on Improved Point Cloud Registration Methods},
  author={Xu, Xiaomei and others},
  booktitle={CASE},
  year={2024},
  organization={IEEE}
}

@article{wang2021trajectory,
  title={Trajectory planning and optimization for robotic machining based on measured point cloud},
  author={Wang, Gang and others},
  journal={IEEE Transactions on robotics},
  volume={38},
  year={2021},
  publisher={IEEE}
}

@article{wen2022uniform,
  title={Uniform coverage tool path generation for robotic surface finishing of curved surfaces},
  author={Wen, Yalun and others},
  journal={IEEE RA-L},
  volume={7},
  year={2022},
  publisher={IEEE}
}

@inproceedings{xia2023human,
  title={Human-Robot Collaboration for Unknown Flexible Surface Exploration and Treatment Based on Mesh Iterative Learning Control},
  author={Xia, Jingkang and others},
  booktitle={IROS},
  year={2023},
  organization={IEEE}
}

@article{schneyer2023segmentation,
  title={Segmentation and coverage planning of freeform geometries for robotic surface finishing},
  author={Schneyer, Stefan and others},
  journal={IEEE RA-L},
  year={2023},
  publisher={IEEE}
}

@article{sheng2001cad,
  title={CAD-guided robot motion planning},
  author={Sheng, Weihua and others},
  journal={Industrial Robot: An International Journal},
  volume={28},
  year={2001},
  publisher={MCB UP Ltd}
}

@inproceedings{liu2016industrial,
  title={Industrial robot path planning for polishing applications},
  author={Liu, Jiang and others},
  booktitle={ROBIO},
  year={2016},
  organization={IEEE}
}

@article{liu2018region,
  title={A region-based 3+ 2-axis machining toolpath generation method for freeform surface},
  author={Liu, Xu and others},
  journal={The International Journal of Advanced Manufacturing Technology},
  volume={97},
  year={2018},
  publisher={Springer}
}

@article{xiao2021model,
  title={A model-based trajectory planning method for robotic polishing of complex surfaces},
  author={Xiao, Mubang and others},
  journal={IEEE Transactions on Automation Science and Engineering},
  volume={19},
  year={2021},
  publisher={IEEE}
}

@article{makhataeva2020augmented,
  title={Augmented reality for robotics: a review},
  author={Makhataeva, Zhanat and Varol, Huseyin Atakan},
  journal={Robotics},
  volume={9},
  year={2020},
  publisher={Multidisciplinary Digital Publishing Institute}
}

@inproceedings{fu2023recent,
  title={Recent advancements in augmented reality for robotic applications: A survey},
  author={Fu, Junling and others},
  booktitle={Actuators},
  volume={12},
  year={2023},
  organization={MDPI}
}

@article{porpiglia2019three,
  title={Three-dimensional elastic augmented-reality robot-assisted radical prostatectomy using hyperaccuracy three-dimensional reconstruction technology: a step further in the identification of capsular involvement},
  author={Porpiglia, Francesco and others},
  journal={European urology},
  volume={76},
  year={2019},
  publisher={Elsevier}
}

@article{dogangun2024rampa,
  title={RAMPA: Robotic Augmented Reality for Machine Programming by Demonstration},
  author={Dogangun, Fatih and others},
  journal={IEEE RA-L},
  year={2025},
  publisher={IEEE}
}

@inproceedings{wang2024eve,
  title={EVE: Enabling Anyone to Train Robots using Augmented Reality},
  author={Wang, Jun and others},
  booktitle={37th Annual ACM Symposium on User Interface Software and Technology},
  pages={1--13},
  year={2024}
}

@article{hoang2022arviz,
  title={Arviz: An augmented reality-enabled visualization platform for ros applications},
  author={Hoang, Khoa C and others},
  journal={IEEE RA-M},
  volume={29},
  year={2022},
  publisher={IEEE}
}

@article{ong2020ar,
  title={AR-assisted robot welding programming},
  author={Ong, SK and others},
  journal={Advances in Manufacturing},
  volume={8},
  year={2020},
  publisher={Springer}
}

@article{tavares2019collaborative,
  title={Collaborative welding system using BIM for robotic reprogramming and spatial augmented reality},
  author={Tavares, Pedro and others},
  journal={Automation in Construction},
  volume={106},
  year={2019},
  publisher={Elsevier}
}

@article{zhang2023interactive,
  title={Interactive Fabrication: Exploration of Human-machine Collaboration Methods for AR Assisted Robotic Fabrication},
  author={Zhang, Weichen},
  year={2023},
  journal={Politecnico Milano}
}

@inproceedings{de2019intuitive,
  title={An Intuitive augmented reality interface for task scheduling, monitoring, and work performance improvement in human-robot collaboration},
  author={De Franco, Alessandro and others},
  booktitle={IWOBI},
  year={2019},
  organization={IEEE}
}

@article{ciccarelli2024advancing,
  title={Advancing human-robot collaboration in handcrafted manufacturing: cobot-assisted polishing design boosted by virtual reality and human-in-the-loop},
  author={Ciccarelli, Marianna and others},
  journal={The International Journal of Advanced Manufacturing Technology},
  pages={1--16},
  year={2024},
  publisher={Springer}
}

@inproceedings{quintero2018robot,
  title={Robot programming through augmented trajectories in augmented reality},
  author={Quintero, Camilo Perez and others},
  booktitle={IROS},
  year={2018},
  organization={IEEE}
}

@inproceedings{stadler2016augmented,
  title={Augmented reality for industrial robot programmers: Workload analysis for task-based, augmented reality-supported robot control},
  author={Stadler, Susanne and others},
  booktitle={RO-MAN},
  year={2016},
  organization={IEEE}
}

@article{martinez2022ransac,
  title={Ransac for robotic applications: A survey},
  author={Mart{\'\i}nez-Otzeta, Jos{\'e} Mar{\'\i}a and others},
  journal={Sensors},
  volume={23},
  year={2022},
  publisher={MDPI}
}

@article{he2020review,
  title={A review of monocular visual odometry},
  author={He, Ming and others},
  journal={The Visual Computer},
  volume={36},
  year={2020},
  publisher={Springer}
}

@inproceedings{kazhdan2006poisson,
  title={Poisson surface reconstruction},
  author={Kazhdan, Michael and others},
  booktitle={Eurographics symposium on Geometry processing},
  volume={7},
  year={2006}
}

@article{schrepp2017construction,
  title={Construction of a benchmark for the user experience questionnaire (UEQ)},
  author={Schrepp, Martin and others},
  journal={International Journal of Interactive Multimedia and Artificial Intelligence},
  volume={4},
  year={2017},
  publisher={Universidad Internacional de La Rioja}
}

@inproceedings{laugwitz2008construction,
  title={Construction and evaluation of a user experience questionnaire},
  author={Laugwitz, Bettina and others},
  booktitle={Symposium of the Austrian HCI and usability engineering group},
  year={2008},
  organization={Springer}
}

@article{hart1988development,
  title={Development of NASA-TLX (Task Load Index): Results of empirical and theoretical research},
  author={Hart, SG},
  journal={Human mental workload},
  year={1988}
}

@article{hurtienne2010quesi,
  title={QUESI—A questionnaire for measuring the subjective consequences of intuitive use},
  author={Hurtienne, J{\"o}rn and Naumann, Anja},
  journal={Interdisciplinary College},
  volume={536},
  year={2010}
}
